\documentclass{article}

\usepackage[preprint]{neurips_2024}

\usepackage[utf8]{inputenc} 
\usepackage[T1]{fontenc}    
\usepackage[colorlinks,linkcolor=blue,anchorcolor=blue,citecolor=blue]{hyperref}
\usepackage{url}            
\usepackage{booktabs}       
\usepackage{amsfonts}       
\usepackage{nicefrac}       
\usepackage{microtype}      
\usepackage{xcolor}         
\usepackage{graphicx,subfigure}
\usepackage{amsmath}
\usepackage{amsfonts}
\usepackage{mathrsfs}
\usepackage{cases}

\usepackage{arydshln}

\usepackage{multicol}
\usepackage{multirow}
\usepackage{threeparttable}

\usepackage[lined,boxed,ruled,noend]{algorithm2e}

\title{Faithful Density-Peaks Clustering via Matrix Computations on MPI Parallelization System}

\author{%
  Ji Xu \thanks{Corresponding author.},\quad Tianlong Xiao,\quad Jinye Yang,\quad Panpan Zhu \\
  State Key Laboratory of Public Big Data\\
  Guizhou University\\
  Guiyang, China 550025 \\
  \texttt{jixu@gzu.edu.cn} \\
}

\begin{document}

\maketitle

\begin{abstract}
 Density peaks clustering (DP) has the ability of detecting clusters of arbitrary shape and clustering non-Euclidean space data, but its quadratic complexity in both computing and storage makes it difficult to scale for big data. Various approaches have been proposed in this regard, including MapReduce based distribution computing, multi-core parallelism, presentation transformation (e.g., $k$d-tree, Z-value), granular computing, and so forth. However, most of these existing methods face two limitations. One is their target datasets are mostly constrained to be in Euclidian space, the other is they emphasize only on local neighbors while ignoring global data distribution due to restriction to cut-off kernel when computing density. To address the two issues, we present a faithful and parallel DP method that makes use of two types of vector-like distance matrices and an inverse leading-node-finding policy. The method is implemented on a message passing interface (MPI) system. Extensive experiments showed that our method is capable of clustering non-Euclidean data such as in community detection, while outperforming the state-of-the-art counterpart methods in accuracy when clustering large Euclidean data. Our code is publicly available at https://github.com/alanxuji/FaithPDP.
\end{abstract}

\section{Introduction}
%
%
%
%
As a fundamental technique to depict the data distribution unsupervisedly, clustering relates closely to various subfields of machine learning such as semi-supervised transductive learning \cite{zhou2007spectral}, \cite{fu2015transductive}, contrastive learning \cite{li2021contrastive}, \cite{sharma2020clustering}, domain-adaption \cite{deng2021joint}, \cite{liang2019aggregating}, and so on. In these fields, the similarity of data within a cluster is very helpful for promoting the efficiency and accuracy of a given learning model. Clustering itself can be used directly in many applications as well, for example, in image segmentation \cite{coleman1979image}, \cite{dong2005color}, \cite{tang2020fuzzy}, community discovery \cite{wang2014neiwalk}, \cite{niu2023overlapping} and recommendation system \cite{farseev2017cross}, \cite{cui2020personalized}. There have been many clustering methods, among which density-peak clustering (DP) \cite{rodriguez2014clustering} is very successful and influential due to its ability to detect clusters of arbitrary shapes and non-iterative computing procedure.

However, there is a bottleneck in the original DP, that is, it requires $O(n^2)$ complexity to compute and store the distance matrix, which prevent DP from directly scaling for big data analytics. Many excellent research works have been put-forward to address this issue that can be categorized into three streams: a) utilizing data structure (utilizing a data-structure \cite{amagata2023efficient} or data encoding \cite{lu2022distributed}), b) modifying computing architecture (Map-reduce distributed or multi-core parallelism) \cite{zhang2016efficient}, \cite{amagata2021fast}, and c) employing new computing methodology (e.g., granular computing) \cite{cheng2023fast}. For the data structure category,

These methods effectively endow DP the ability to process large scale data, but leave two main issues unaddressed. The first is because of relying on data organization, approximate encoding, or data splitting and merging, they \textbf{lost the ability of capturing the local sophisticated structure and global data distribution} (which is the first core advantage of DP). Fig. \ref{fig:SprialsComparison} provides an illustrative example. The second is that most of them typically \textbf{constrained their target data to lie in Euclidean space}, and lost the ability of taking a distance as input to perform clustering (which is the second core advantage of DP).

To solve the aforementioned weaknesses of the existing methods that scales DP, we propose in this paper a faithful parallel DP (FaithPDP) that achieves the identical intermediate results and final clustering result to original plain DP, while keeping the memory consumption linear to data size. Because the provable faithfulness of the proposed parallelism of DP, FaithPDP keeps the two core advantages of vanilla DP, that is, it can capture the micro subtle structure of data thus achieves high accuracy and it has the ability to clustering non-Euclidean data (e.g., clustering nodes in a social network). The contributions of our method include three aspects:

\begin{enumerate}
\item The proposed method reduced the computing time and memory occupation to enable DP to clustering big data with the intermediate results and final result being identical to original DP. Our method achieves pseudo-linear time complexity and linear space complexity.

\item FaithPDP keeps the same ability to cluster non-Euclidean data (e.g., social network node cluster) as vanilla DP, while being able to accommodate large scale data.

\item Theoretical analysis and extensive experiments through massage passing interface (MPI) \cite{nielsen2016introduction} show that our method achieves superior accuracy while keeping competitive efficiency when dealing with either Euclidean or graph data.

\end{enumerate}

\section{Related Works}

There have been a number of works aiming to scale DP to cluster massive data, which can be categorized as exact and approximate.  By ``exact", it means the method can generate exactly the same intermediate and final results as DP with integer cut-off kernel, rather than with Gaussian kernel that leads to real density.

\subsection{Exact Methods}
\textbf{Data structure.} Ex-DPC \cite{amagata2023efficient} employs \emph{k}d-tree structure to facilitate fast computing of local density $\rho$ and depending distance $\delta$.
All methods based on \emph{k}d-tree use cut-off kernel to compute density, which causes integer density and subsequent random sorting of the objects with the same density. Besides, they calculate the position of each data point through iterative partitioning on each dimension, thus prefer data with low dimensionality.

\textbf{Computing architecture.} The earliest effort to scaling DP for big data is taken by Zhang \emph{et. al.} \cite{zhang2016efficient} introducing MapReduce framework to distributedly clustering data. In their paper, locality sensitive hash (LSH) is utilized to avoid high communication and computation overload of Basic-DDP. Later, multi-core based parallelization is introduce to accelerate DP \cite{amagata2021fast}, which is the pioneering work based on \emph{k}d-tree and recently improved and consummated as in \cite{amagata2023efficient}.

\subsection{Approximate Methods}
Ex-DPC++ \cite{amagata2023efficient} uses \emph{k}d-tree and cover-tree (another data structure to accelerate nearest neighbor query in any metric space) \cite{beygelzimer2006cover}, to expand the application scenario of Ex-DPC. CFSFDP+A uses a \emph{k}-means clustering to obtain an initial partition of the data, and reduces the workload of computation of density and $\delta$-distance (called separation therein) based on the partition \cite{bai2017fast}. Similarly, GB-DP \cite{cheng2023fast} employs \emph{k}-means (usually \emph{k} is set to be 2) to construct granular balls, and perform DP clustering based on the granular balls rather than original data points, hence achieve considerable acceleration in computing.

\section{Preliminaries}
\noindent \textbf{Density Peaks Clustering (DP).} DP consists of three major steps, aiming to compute four key vectors (namely, local density $\boldsymbol \rho$, depending node $\boldsymbol \mu$, depending distance $\boldsymbol \delta$, and center potential $\boldsymbol \gamma$) based on the pair-wise distance matrix $D=\{d_{ij}\}_{i, j=1}^{n}$. It dose not matter whether $D$ is computed from a dataset $\mathcal{X}$ in Euclidean space or from a graph $\mathcal{G}$. The goal is to assign a cluster label $y_i \in \{1,...,C\}$ to ecah $x_i \in \mathcal{X}$ or a node $v_i$  in graph $ \mathcal{G}$ . We use $\mathcal{Y}=(y_1,...,y_n)$ to denote all the clustering labels and ${\mathcal C}$ the set of centers. DP consists of three steps as follows.\\
(1)	Local density $\boldsymbol{\rho}=(\rho_1,...\rho_n)$, where $\rho_i= \sum\nolimits_{j \ne i} {\exp \big( - d_{ij}/d_c\big) ^2}$.\\
(2) Depending node $\boldsymbol{\mu}=(\mu_1,...\mu_n)$ and depending distance $\boldsymbol{\delta}=(\delta_1,...\delta_n)$ \footnote{These also named \emph{dependent points} and \emph{dependent distance} in \cite{amagata2023efficient}. In addition, they are also named as \emph{leading node} and $\delta$\emph{-distance} in \cite{xu2016denpehc}}, where $\mu_i  = \mathop {\arg \min }\limits_{\boldsymbol x_j} \{ {d_{ij}}|{\rho _j} > {\rho _i}\}$ and ${\delta _i} = \min \{ {d_{ij}}|{\rho _j} > {\rho _i}\}$.\\
(3) Center potential $\boldsymbol{\gamma}=\boldsymbol{\rho} \odot \boldsymbol{\delta}$, where $\odot$ denotes Hadamard product.\\
The clustering process is accomplished by adaptive selecting the centers featured with extraordinarily large elements in $\gamma$ and assign the rest of the data points to each cluster according to a chain rule, i.e.,
\begin{equation}\label{eq:AssignLabel}
   y_i = \left\{ \begin{gathered}
  j, if \; x_i = {\mathcal C}_ j; \hfill \\
  y_{\mu_i}, \; otherwise.  \hfill \\
\end{gathered}  \right.
\end{equation}

\noindent \textbf{Accelerating DP via matrix computing.}  Matrix based computation of $D$ has been proposed to replace the pair-wise distance computation, for either Euclidean distance or cosine similarity as in Eq. (\ref{eq:MatDistCompute})
\begin{equation}\label{eq:MatDistCompute}
    D^{Eu}(\mathcal{P}_i,\mathcal{X}) = sqrt(\mathcal{P}_i^{\circ2}{\boldsymbol{1}_{d \times n}} + {\boldsymbol{1}_{m \times d}}  (\mathcal{X}^\top)^{\circ2} - 2\mathcal{P}_i  {\mathcal{X}^\top}),
\end{equation}
or  Eq. (\ref{eq:CosDistMatCompute}) \cite{xu2021lapoleaf}
 \begin{equation}\label{eq:CosDistMatCompute}
   D^{cos}(\mathcal{P}_i,\mathcal{X}) = \boldsymbol{1}_{m \times n}- {\mathcal{P}  {\mathcal{X}^\top}}\circ/{sqrt(\mathcal{P}_i^{\circ2} \boldsymbol{1}_{d \times d}  (\mathcal{X}^\top)^{\circ2} )},
\end{equation}

\begin{figure*}[t]
  \centering
  \includegraphics[width=\textwidth]{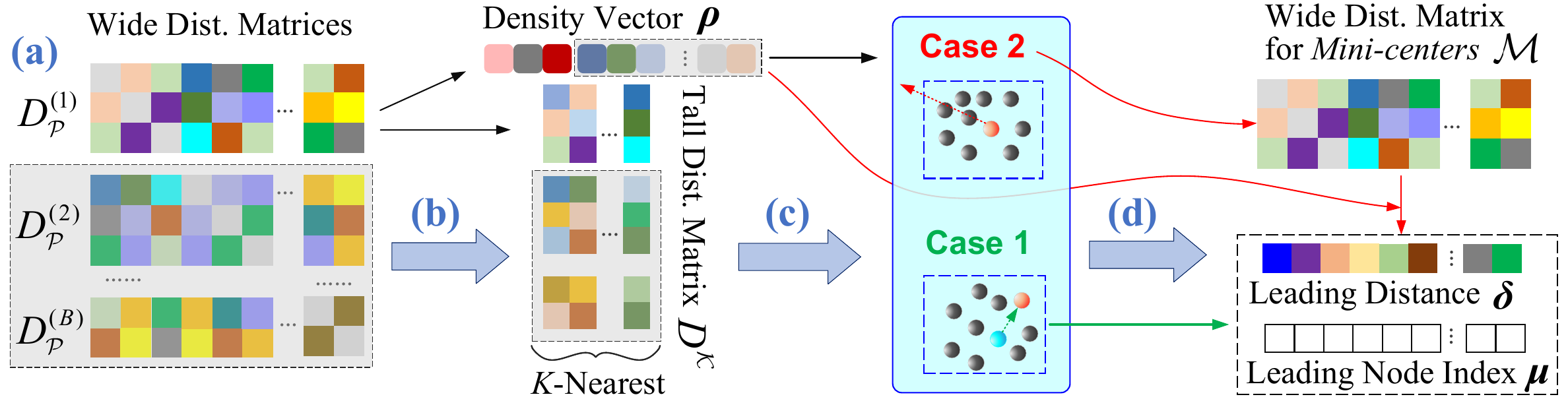}
  \footnotesize
  \caption{The steps of computing the key vectors in FaithPDP. (a) Compute the distances of a part of the samples against all data using matrix formulation (so the computations of tall distance matrix ${D}^{\mathcal{K}}$ and $\boldsymbol \rho$ are parallelizable). (b) The segment of density vector and the block of tall distance matrix are computed based on a wide distance matrix. (c) Most of the depending data and depending distance are computed via inverse density-distance condition, and those data points that cannot find depending data are identified as mini centers for further processing (parallelizable). (d) Compute the wide distance matrix for mini centers to decide the remaining $\mu$ and $\delta$ (centralized). So far, all the three key vectors for DP are worked out.}\label{fig:FaithPDPDiagram}
\end{figure*}

where $\mathcal{P}_i$  is a subset of $\mathcal{X}$ (or the embedding of nodes in graph  $\mathcal{G}$) consists of samples in the form of row vector, $\circ2$ and $\circ/$ are the operators of element-wise square and element-wise division, respectively. We omit the superscript $Eu$ or $cos$ afterwards and denote $D(\mathcal{P}_i,\mathcal{X})$  as $D_{\mathcal{P}}^{(i)}$ for short. We use Euclidean distance as default unless specified otherwise. Consequently, the local density vector segments of a subset samples $P$ can be computed as in Eq. (\ref{eq:MatRhoCompute}),
\begin{equation}\label{eq:MatRhoCompute}
\boldsymbol{\rho}_{\mathcal{P}}^{(i)}= {\rm sum}\Big(\circ\exp\big((D_{\mathcal{P}}^{(i)})/d_c)^{\circ2}\big), {\rm axis}=``{\rm row}" \Big),
\end{equation}
in which all the elements in matrix $D_{\mathcal{P}}^{(i)}$ are first divided by $d_c$, then take Hadamard square and Hadamard exponential. Finally, the elements in each row are summed to get the local density of corresponding sample.\\

\noindent\textbf{Parallelism via Message-Passing Interface.} Message Passing Interface (MPI) is a standardized and portable message-passing system designed to implement parallel computing, which is established since 1994 and still under active updating nowadays \cite{mpi41}. The basic functions of MPI that facilitate parallel computing include message sending/receiving, gathering/broadcasting, and so forth. Early MPI supports mainstream programming languages (such as C/C++, Fortran), and later MPI for Python was developed \cite{dalcin2005mpi, dalcin2021mpi4py}. For efficiency of development, we choose MPI4py in our implementation.

\section{Faithful Parallel DPC}

To make our method easier to follow, we depict our computing flowchart in Fig. \ref{fig:FaithPDPDiagram} and first describe our method of sequential version in Algorithm \ref{alg:SeqFaithDP}, then a final faithful parallel DP (FaithPDP) is described in Algorithm .

\begin{algorithm}[h]
\caption{Sequential FaithDP }
\label{alg:SeqFaithDP}
\LinesNumbered
\KwIn{A dataset $\mathcal{X} \in \mathbb{R}^{d}$ or a graph  $\mathcal{G} =(V, E)$, $K$ (number of nearest neighbors seeking to identify depending data point).}
\KwOut{ Cluster labels $\mathcal {Y}$ for instances (in $\mathcal {X}$) or nodes ($V$ of $\mathcal {G}$)}
\tcp{\small Stage 1: Computing density vector segments and tall distance matrices.}
Divide $\mathcal {X}$ or $V$ into $B$ batches;\\
\For {each $\mathcal{P}_i$}
{Compute $D_{\mathcal{P}}^{(i)}$ using Eq. (\ref{eq:MatDistCompute}) or Eq. (\ref{eq:CosDistMatCompute});\\
 Compute $\boldsymbol{\rho}_{\mathcal{P}}^{(i)}$ using Eq. (\ref{eq:MatRhoCompute});\\
$[{D}^{\mathcal{K}}_{\mathcal{P}_i},\mathcal{N}_{\mathcal{P}}^{(i)} ] \leftarrow$  Find $k$ nearest neighbors for samples in $\mathcal{P}_i$;\\
}
Collect all $\boldsymbol{\rho}_{\mathcal{P}}^{(i)}$ , ${D}^{\mathcal{K}}_{\mathcal{P}_i}$ , and $\mathcal{N}_{\mathcal{P}}^{(i)}$ to produce entire $\boldsymbol{\rho}$, ${D}^{\mathcal{K}}$, and $\mathcal{N}$\\
\tcp{\small Stage 2: Computing leading nodes vector and $\delta$-distance vector;}
Initialize mini centers ${\mathcal M} \leftarrow \emptyset $;\\
\For {each $\mathcal{P}_i$}
 {\For {each  $x_h \in \mathcal{P}_i$}
  {
    $j$ = find global index of $h$ ;\\
    $\tilde{\mathcal{N}}_{\mathcal{P}}^{(i)}[h,:] \leftarrow$ sort the elements in $\mathcal{N}_{\mathcal{P}}^{(i)}[h,:]$ by  distances in ${D}^{\mathcal{K}}_{\mathcal{P}_i}[h,:]$;\\

    \For {k<K}{
    \If {${\rho}_j < {\rho}_q $ for  $q = \tilde{\mathcal{N}}_{\mathcal{P}}^{(i)}[h,k]$ }
     {
      ${\mu_j} \leftarrow q $ ,  ${\delta}_j \leftarrow {D}^{\mathcal{K}}_{\mathcal{P}_i}[h,k]$; \\
      \bf{break};\\
     }
     }
     \If{${\mu_j} $ not found}
     {
       ${\mathcal M} \leftarrow {\mathcal M} \cup \{x_j\}$;\\
     }

   }
}
Compute ${\delta}_l$ and ${\mu_l}$ for each $x_l \in { \mathcal M}$;

\tcp{\small Stage 3: Getting final clustering result.}
Decide number of clusters $C$;\\
$\boldsymbol{\gamma} \leftarrow \boldsymbol{\rho} \odot \boldsymbol{\delta}$;\\
Centers set $\mathcal {C} \leftarrow \mathcal {X}[{\rm IdxOf}(topC(\boldsymbol{\gamma}))]$ ;\\
Get $\mathcal{Y}$ using Eq. (\ref{eq:AssignLabel});\\

\Return $\mathcal{Y}$.
\end{algorithm}

Algorithm \ref{alg:FaithPDP} described the faithful and parallel DP based on MPI system.

\begin{algorithm}[h]
\caption{FaithPDP }
\label{alg:FaithPDP}
\LinesNumbered
\KwIn{A dataset $\mathcal{X} \in \mathbb{R}^{d}$ or a graph  $\mathcal{G} =(V, E)$, $K$ (number of nearest neighbors), $R$ (number of parallel procedures).}
\KwOut{ Cluster labels $\mathcal {Y}$.}
\tcp{\small Stage 1: Computing density vector segments and tall matrices \emph{in parallel}.}
Create initial  $\boldsymbol{\rho}$, ${D}^{\mathcal{K}}$, and $\mathcal{N}$;\\
Divide $\mathcal {X}$ into $R$ groups ( each to feed into a procedure );\\
\For {each procedure $Rank_j$}
{
  Line 1 to Line 7 in Alg.  \ref{alg:SeqFaithDP} to compute $\boldsymbol{\rho}^{(j)}$, ${D}^{\mathcal{K}}_{j}$ and $\mathcal{N}^{(j)}$;\\
  Send $\boldsymbol{\rho}^{(j)}$, ${D}^{\mathcal{K}}_{j}$ and $\mathcal{N}^{(j)}$ to $Rank_0$;\\
  \If{$Rank_j$ == $Rank_0$}
  {collect and store $\boldsymbol{\rho}^{j}$, ${D}^{\mathcal{K}}_{j}$ and $\mathcal{N}^{j}$;\\}
}

\tcp{\small Stage 2: Computing leading nodes and leading distances \emph{in parallel};}
Initialize $\mathcal{M} = \emptyset$; $\boldsymbol{\mu} = -\boldsymbol{1}$; $\boldsymbol{\delta} = \boldsymbol{0}$;\\
\For {each procedure $Rank_j$}
{
  Line 8 to Line 17 in Alg. \ref{alg:SeqFaithDP} to compute segments $\boldsymbol{\mu}^{(j)}$ and $\boldsymbol{\delta}^{(j)}$;\\
  Send $\boldsymbol{\mu}^{(j)}$ and $\boldsymbol{\delta}^{(j)}$ to $Rank_0$;\\
  \If{$Rank_j$ == $Rank_0$}
  {collect $\boldsymbol{\mu}^{(j)}$ and $\boldsymbol{\delta}^{(j)}$; \\
   $\mathcal{M} \leftarrow$ IdxOf($\boldsymbol{\mu}$==-1);\\
   $D^\mathcal{M} \leftarrow$ wide distance between $\mathcal{M}$ and $\mathcal{X}$;\\
    Use $D^\mathcal{M}$  and  $\boldsymbol \rho$ to determine $\mu$s and $ \delta$s for samples in $\mathcal{M}$;\\
  }
}

\tcp{\small Stage 3: Computing final clustering result \emph{sequentially}.}
Line 19 to Line 21 in Alg. \ref{alg:SeqFaithDP} to get clustering result $\mathcal{Y}$;\\

\Return $\mathcal{Y}$.
\end{algorithm}

\section{Experiments}
All the experiments are conducted on two Dell T7920 workstations \footnote{Each equipped with two Xeon 6230R CPUs, 64G DDR4 Memory, an NVIDIA GForce 3090 GPU, 512G SSD + 4T HDD Hard Disk.}, Python 3.9 and mpi4py 3.1.5 running in a Conda virtual environment configured on the operating system of Ubuntu 22.04.
\subsection{Design purpose}
Our experiment is designed to answer the following questions.\\
\textbf{Q1}: What the difference between FaithPDP and the existing scaling-up methods of DP?\\
\textbf{Q2}: How is the performance improvement of FaithPDP over other competing models on learned representation of real data?\\
\textbf{Q3}: How does FaithPDP perform on high-dimensional Euclidean data?\\
\textbf{Q4}: Is FaithPDP perform efficiently while being accurate?\\

\subsection{Datasets and baseline models}
To answer the research questions Q1\textasciitilde Q5, the chosen datasets vary from medium scale to huge scale, from Euclidean to graph, from synthetic to real, and from low-dimensional to high-dimensional. The information of the datasets used in our empirical evaluations is listed in Table \ref{tab:FaithPDPDatasets}.
\begin{table*}
  \centering
    \renewcommand{\arraystretch}{1.3}

  \caption{Information of the datasets in the experiments.  }\label{tab:FaithPDPDatasets}
  \resizebox{0.9\textwidth}{!}{
  \begin{tabular}{ccccccc}
    \toprule
 SN&Dataset& Type & Source & Data size $^{\dag}$ & \# Clusters &Purpose\\
 \hline
 DS1&5Spriral50k&Euclidean& Synthetic &(52,834; 2)&5 &Q1\\
 DS2&5Spiral500k&Euclidean& Synthetic &(528,320; 2)&5 &Q1, Q4\\
DS3& Pamap2\_sub1&Euclidean& Real &(7,528; 8,109)&15 &Q2, Q3\\
DS4&Opportunity\_s2ADL3 &Euclidean& Real &(34,232; 243)&4 &Q2, Q3\\
DS5&MINIST159& Euclidean& Real &(18,112; 784)&3 &Q2, Q3\\
DS6&MINIST159\_emb& Euclidean& Real &(18,112; 2)&3 &Q2\\
DS7&MINIST\_tr& Euclidean& Real &(60,000; 784)&10 &Q1, Q3\\
DS8&Email\_emb& Euclidean& Real &(1,005; 64)&41 &Q2\\
  \hline

  \end{tabular}}
    \begin{tablenotes}
     \item[1] {\footnotesize $\dag$: ($\cdot$; $\cdot$) denotes the number of samples and dimensions and <$\cdot$; $\cdot$> the number of nodes and edges.}
    \end{tablenotes}
\end{table*}

We evaluated our FaithPDP against six state-of-the-art methods. They are \\
1) \textbf{Index-List} \cite{rasool2022index}: a method that construct a list of nearest neighbors for fast density computing;\\
2) \textbf{RN-List}: an approximate alternative of Index-List;\\
3) \textbf{GB-DP} \cite{cheng2023fast}: changes the clustering objects from finest-grained data points to granular balls built with hierarchical\emph{ k}-means splitting;\\
4)\textbf{ Ex-DPC} \cite{amagata2021fast, amagata2023efficient}: fast density and leading point computation by restricting search field in a \emph{k}d-tree;\\
5) \textbf{Z-Value} \cite{lu2022distributed}: fast DP through encoding the samples into a integer space;\\
6) \textbf{Ex-DPC}++ \cite{amagata2023efficient}: Improve Ex-DPC further with another data structure named cover-tree.\\

\subsection{Performance comparison and discussion}

\noindent\textbf{Performance on accuracy.} The accuracy is evaluated under the metrics of normalized mutual information (NMI) \cite{strehl2002cluster} and adjusted Rand index (ARI) \cite{hubert1985comparing}. Table shows the detailed accuracies achieved by FaithPDP and the competing models on various datasets. For human activity recognition (HAR) data DS3 and DS4, we use empirical cumulative distribution function (ECDF) to extract meaningful features \cite{plotz2011feature}.

\begin{figure*}[t]
  \centering
  \includegraphics[width=0.68\textwidth]{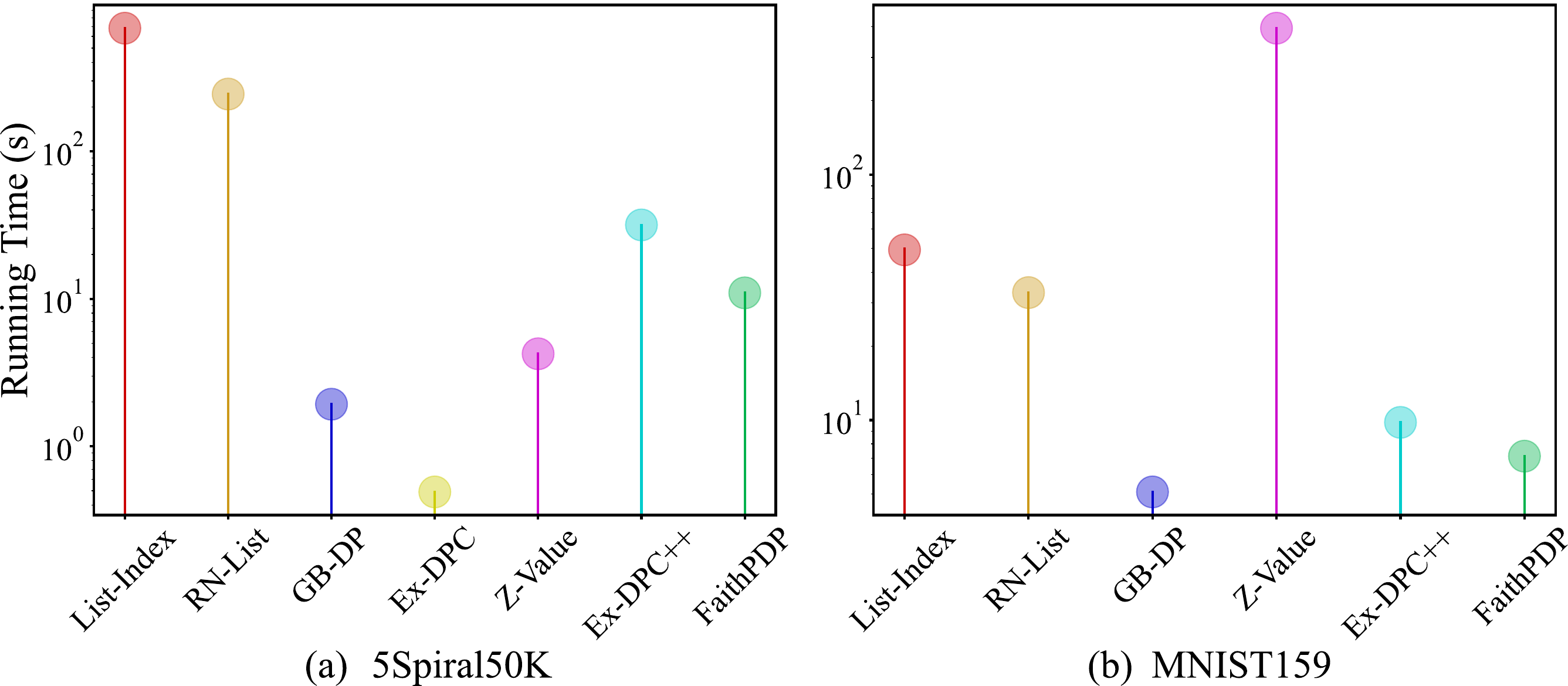}
  \footnotesize
  \caption{Running time comparisons.  }\label{fig:RunningTime}
\end{figure*}

\begin{table*}
  \centering
    \renewcommand{\arraystretch}{1.3}

  \caption{Information of the datasets in the experiments.  }\label{tab:FaithPDPAcc}
  \resizebox{0.9\textwidth}{!}{
  \begin{tabular}{ccccccccc}
    \toprule

   &  &Index-List &RN-List &GB-DP & Ex-DPC &Z-value &Ex-DPC++ & FaithPDP\\
 \hline

 \multirow {2}{*} {DS1} &NMI    &\emph{\underline{0.8521 }} &\emph{\underline{0.8521 }} &0.4563 &0.7443  &0.0113  &0.6849 &\textbf{1.0000} \\
    &ARI    &\emph{\underline{0.8206 }} &\emph{\underline{0.8206}} &0.3510  &0.5701  & 0.4271 &0.1082 & \textbf{1.0000} \\

\hdashline[0.5pt/3pt]
 \multirow {2}{*} {DS2} &NMI    & \multirow {2}{*} {OOM}  & \multirow {2}{*} {OOM}  & 0.3673 &0.7445  &0.0137  &\emph{\underline{0.9868}} &\textbf{1.0000} \\
    &ARI   &  & &0.2215 &0.5703 & 0.0058 &\emph{\underline{0.9891}} & \textbf{1.0000} \\

\hdashline[0.5pt/3pt]
 \multirow {2}{*} {DS3} &NMI    &\emph{\underline{0.3780}}  &0.3500  & 0.2021 &\multirow {2}{*} {--}   &\multirow {2}{*} {VOF}   &\multirow {2}{*} {--}  &\textbf{0.4038} \\
    &ARI   &\emph{\underline{0.1470}}  &0.1343 & 0.0632 & &   && \textbf{0.1641} \\

\hdashline[0.5pt/3pt]
 \multirow {2}{*} {DS4} &NMI    &0.3740  &0.4128 & 0.2134 & \textbf{0.4573}  &\multirow {2}{*} {VOF}   &0.2370  &\emph{\underline{0.4480}} \\
    &ARI   &0.1977 &0.2500 & 0.1160&\textbf{0.2704} &   &0.1294& \emph{\underline{0.2672}} \\

   \hdashline[0.5pt/3pt]

 \multirow {2}{*} {DS5} &NMI    & 0.3681  &\emph{\underline{0.3693 }} & 0.1873 &\multirow {2}{*} {--}   &\multirow {2}{*} {VOF}  &\multirow {2}{*} {--}  &\textbf{0.3794} \\
    &ARI   &0.2368  &\emph{\underline{0.2572}} & 0.0764 & &  & & \textbf{0.2676} \\
\hdashline[0.5pt/3pt]

 \multirow {2}{*} {DS6} &NMI    &\emph{\underline{ 0.9494 }} &0.5545 &0.6649 &\multirow {2}{*} {--}   &0.5727  &0.8690 &\textbf{0.9507} \\
    &ARI   &\emph{\underline{0.9734}}  &0.4901 &0.5779 & &0.5565  &0.8981 & \textbf{0.9741} \\

 \multirow {2}{*} {DS7} &NMI    & \multirow {2}{*} {OOM} & \multirow {2}{*} {OOM} & \emph{\underline{0.4384}} &\multirow {2}{*} {OOM}  &\multirow {2}{*} {VOF}  &\multirow {2}{*} {OOM} &\textbf{0.4516} \\
    &ARI   & & &\emph{\underline{0.2489}} & &  & & \textbf{0.2667} \\

\hdashline[0.5pt/3pt]
 \multirow {2}{*} {DS8} &NMI    & 0.1536  &0.2039 &0.6419 &\multirow {2}{*} {--}  &\multirow {2}{*} {VOF}   &\emph{\underline{0.6784}} &\textbf{0.6972} \\
    &ARI   &0.0017  &0.0009 &0.3051 & &  &\emph{\underline{0.3573}} & \textbf{0.4228} \\

 \bottomrule
  \end{tabular}
  }

\end{table*}

\begin{figure*}[b]
  \centering
  \includegraphics[width=0.9\textwidth]{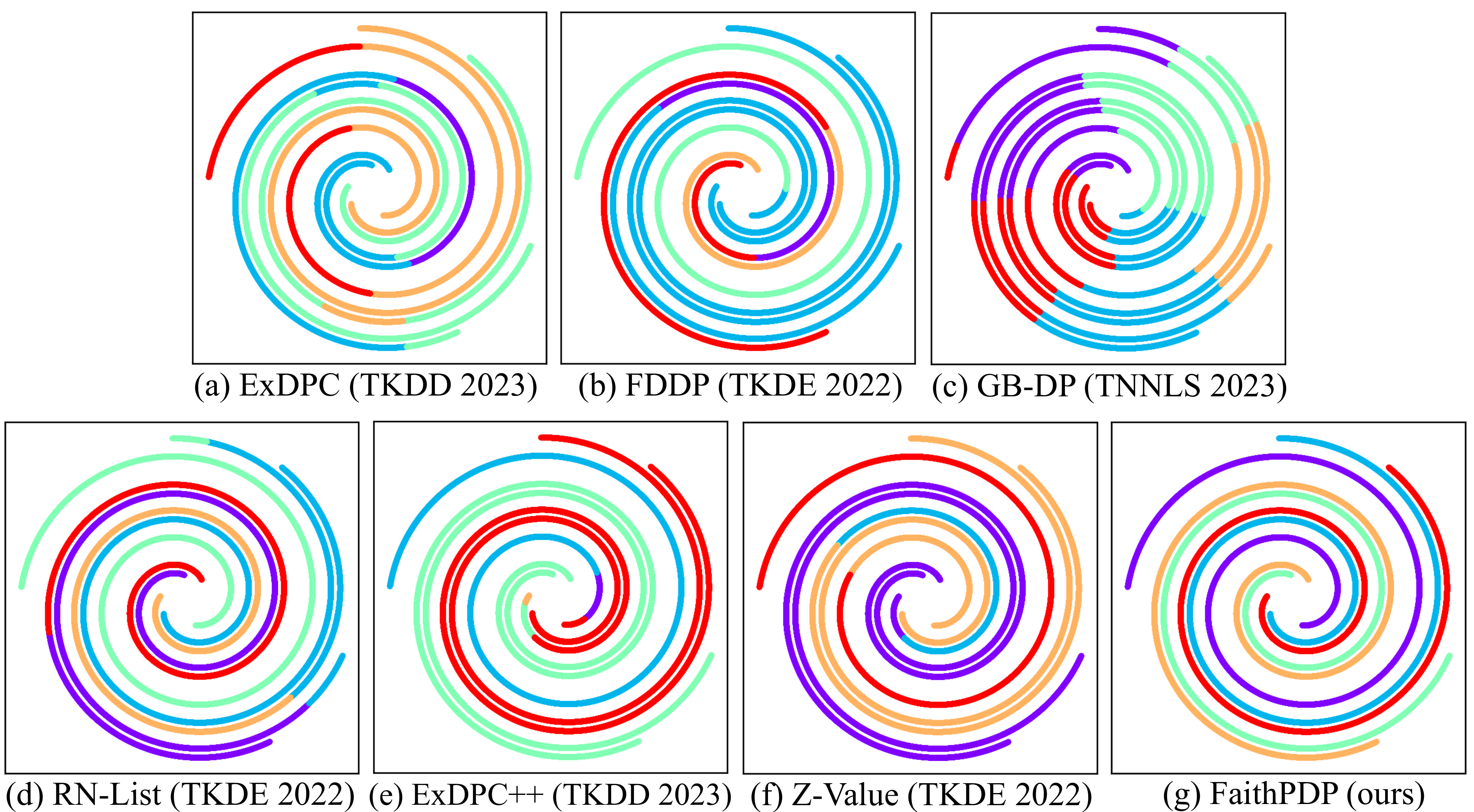}
  \footnotesize
  \caption{The empirical evaluations of six comparative methods and our proposed method on the 5sprial50K dataset (generated by ``FiveSpiralData.py"). }\label{fig:SprialsComparison}
\end{figure*}

The two synthetic datasets are designed to demonstrate the accuracy and efficiency of FaithPDP. To better recognize the core advantage of FaithPDP, we visualize the clustering results of our method and six state-of-the-art counterparts in Fig. \ref{fig:SprialsComparison}, from which one can find that FaithPDP is the only method that perfectly captures the subtle structure the spirals.

\noindent\textbf{Performance on efficiency.} We report the efficiency performance on three datasets that are representatives of two categories (considering number of samples and dimensionality) of dataset respectively. They are large amount with low dimension (DS1 and DS5) and Large amount with large dimension (DS7). For DS1 and DS5, the running time is visualized in Fig. \ref{fig:RunningTime}, from which one can find FaithPDP achieves comparable time efficiency while generally being more accurate (refer to Table \ref{tab:FaithPDPAcc}). For DS7, only GB-DP and FaithPDP can accomplish the clustering task on our experiment environment, the time assumptions are 37.95 seconds and 40.37 seconds, respectively.

\noindent\textbf{Discussion.} Table \ref{tab:FaithPDPAcc} shows that among eight datasets, FaithPDP wins the first place in accuracy for seven times and the second place once. The key insight is:\\
 a) the exact family and list-index-based methods  rely on cut-off kernel, hence emphasize on local neighbors while neglecting global distribution of entire dataset.\\
 b) the methods combined with \emph{k}-means first preprocess datapoints into spherical micro-clusters, hence lose the ability of DP to trace curve distribution (definitely exists  in real world, although it is less common than spherical).\\
 c) coding based methods (including Z-value and LSH) approximate map the original data into a new space, therefore they will be subjected to sacrifice of accuracy.\\
 d) most existing methods compute the distances between data points in a pair-wise manner, thus causing relatively low efficiency.\\

 FaithPDP addressed the issues mentioned above simultaneously, therefore achieved very promising accuracy and overall good time efficiency (slower only than GB-DP).

\section{Conclusion}
This paper presents a novel approach FaithPDP that takes advantages of both hardware (multi-core architecture of CPU) and modern programming language (Python or Matlab for efficient vector and matrix computation) to achieve clustering result identical to vanilla DP algorithm, while the computing complexity is reduced to pseudo-linear. FaithPDP addressed several issues raised by the existing counterparts simultaneously, a) preference to low-dimensional data by \emph{k}d-tree based methods, b) degrading in accuracy by approximate approaches (\emph{k}-means involved, LSH or Z-valued based), c) restricted to cut-off kernel in computing density by \emph{k}d-tree and Index-list based methods. FaithPDP computing distance matrices in a matrix-computation manner other than pair-wise manner and utilize a inverse distance-density condition to find leading points. These two techniques enabled fast computation of density (compatible with both cut-off kernel and Gaussian kernel) and avoid $O(n^2)$ memory requirement for storing the entire distance matrix, respectively. Extensive experiments show that FaithPDP has superior accuracy and comparable temporal and spacial efficiency against state-of-the-art methods.

\section*{Acknowledgement}
This work has been supported by the National Natural Science Foundation of China under grants 62366008 and 61966005.

\bibliographystyle{unsrt}

\end{document}